\documentclass[10pt,twocolumn,letterpaper]{article}

\usepackage{iccv}
\usepackage{times}
\usepackage{epsfig}
\usepackage{graphicx}
\usepackage{amsmath}
\usepackage{amssymb}

\usepackage{epsfig}
\usepackage{amssymb}
\usepackage{multirow}
\usepackage{bbding}
\usepackage{xspace}
\usepackage{enumerate}
\usepackage{makecell}
\usepackage{caption}
\usepackage{subcaption}
\usepackage[bottom]{footmisc}

\usepackage{algorithm}
\usepackage{algorithmic}
\usepackage{mathrsfs} 
\usepackage{amsmath,bm}
\usepackage{color}
\usepackage{caption}
\usepackage{booktabs}
\usepackage{array}
\usepackage{float}
\usepackage{makecell}

\def\eg{{\em e.g. }}
\def\ie{{\em i.e. }}
\def\etal{{\em et al. }}

\usepackage[pagebackref=true,breaklinks=true,letterpaper=true,colorlinks,bookmarks=false]{hyperref}

\iccvfinalcopy 

\def\iccvPaperID{1594} 

\ificcvfinal\fi
\begin{document}

\title{See Better Before Looking Closer: Weakly Supervised Data Augmentation Network for Fine-Grained Visual Classification}

\author{
Tao Hu$^1$, Honggang Qi$^1$, Qingming Huang$^1$, Yan Lu$^2$\\
$^1$ University of Chinese Academy of Sciences, Beijing\\
$^2$ Microsoft Research Asia, Beijing\\
\textit {Email: hutao16@mails.ucas.ac.cn}
}

\maketitle

\begin{abstract}
Data augmentation is usually adopted to increase the amount of training data, prevent overfitting and improve the performance of deep models. However, in practice, random data augmentation, such as random image cropping, is low-efficiency and might introduce many uncontrolled background noises. In this paper, we propose Weakly Supervised Data Augmentation Network (WS-DAN) to explore the potential of data augmentation. Specifically, for each training image, we first generate attention maps to represent the object's discriminative parts by weakly supervised learning. Next, we augment the image guided by these attention maps, including attention cropping and attention dropping. The proposed WS-DAN improves the classification accuracy in two folds. In the first stage, images can be seen better since more discriminative parts' features will be extracted. In the second stage, attention regions provide accurate location of object, which ensures our model to look at the object closer and further improve the performance. Comprehensive experiments in common fine-grained visual classification datasets show that our WS-DAN surpasses the state-of-the-art methods, which demonstrates its effectiveness.
\end{abstract}

\section{Introduction}
\label{sec:introduction}
Data augmentation is a frequently-used strategy that can improve the generalization of deep models since it increases the number of training data by introducing more data variances. It has been proven to be effective in most of the computer vision tasks, such as object classification, detection, and segmentation. There are various kinds of data augmentation for deep models, including image cropping, rotation and color distortion. Previous works usually choose random data augmentation to pre-process their training data. For instance, random image cropping can generate images with different translations and scales so as to increase the robustness of deep models. However, the cropped regions are randomly sampled and a high percentage of them contain many background noises, which might lower the training efficiency, affect the quality of the extracted features and cancel out its benefits. To improve the efficiency of data augmentation, the model should be aware of spatial information of target objects.

\begin{figure}[t]
    \begin{center}
        \includegraphics[width=0.8\linewidth]{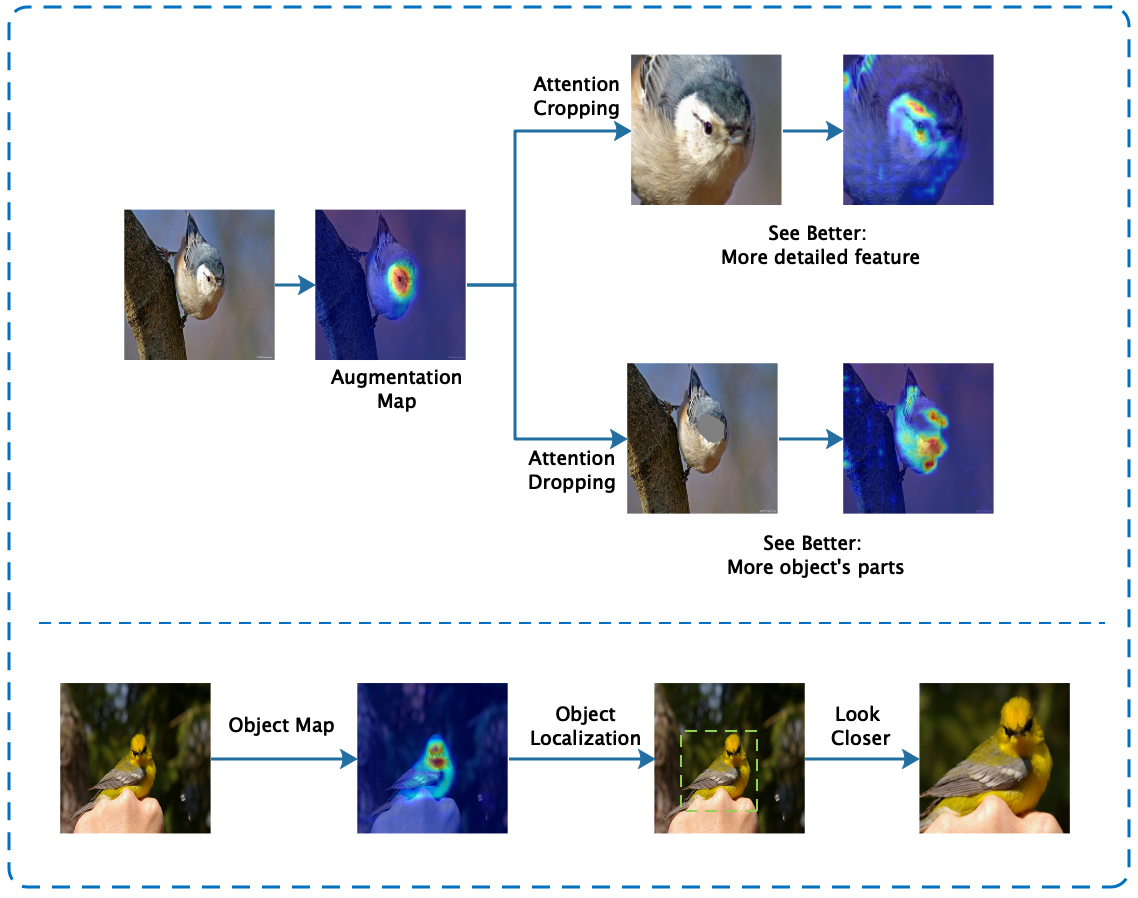}
    \end{center}
    \caption{\small \textbf{See Better}: Attention maps represent discriminative parts of the object. We randomly choose one of the part regions, then drop it to generate more discriminative object's parts or crop it to extract more detailed part feature. \textbf{Look Closer}: The whole object is localized from attention maps and enlarged to further improve the accuracy.}
    \label{fig:attention_augmentation}
\end{figure}

Fine-Grained Visual Classification (FGVC) aims to classify the subordinate-level categories under a basic-level category, such as species of the bird,  model of the car and type of the aircraft. FGVC is challenging because of three main reasons: (1) High intra-class variances. Objects that belong to the same category usually present significantly different poses and viewpoints; (2) Low inter-class variances. Objects that belong to different categories might be very similar apart from some minor differences, \eg the color styles of a bird's head can usually determine its category; (3) Limited training data. Labeling fine-grained categories usually requires specialized knowledge and a large amount of annotation time. Because of the these reasons, it is hard to obtain accurate classification results only by the state-of-the-art coarse-grained Convolutional Neural Networks (CNN), such as VGG~\cite{vgg}, ResNet~\cite{resnet} and Inception~\cite{inception}.

As is pointed out in recent works~\cite{ra-cnn, ma-cnn, mask-cnn}, the key steps of FGVC is extracting more discriminative local features in multiple object's parts. However, object's parts are difficult to be defined and vary from object to object. Moreover, labeling these object's parts requires additional human cost. In this work, we utilize weakly supervised learning to locate the discriminative object's parts only by image-level annotation. Instead of proposing region of interest bounding boxes~\cite{wsddn,ra-cnn}, we represent object's parts or visual pattern by attention maps which are generated by the convolutions. We also propose bilinear attention pooling and attention regularization loss to weakly supervise the attention generation process. Compared with other part localization model~\cite{ma-cnn,mask-cnn}, our model can be more easily to locate a large number of object's parts (more than 10) so as to achieve better performance.

After obtaining the locations of objects' parts, we propose attention-guided data augmentation to effectively augment the training data and solve the above mentioned high intra-class variances and low inter-class variances issues. For different fine-grained categories, objects are usually very similar, apart from few differences. Attention cropping can distinguish them by cropping and resizing the part regions to extract more discriminative local features. For the same fine-grained category, if the model only focuses on few object's parts, it is very likely to predict the wrong category when these parts are occluded because of pose and viewpoint variances. Therefore, it is crucial to extract the local features from different object's parts. Our attention dropping randomly erases one region of objects' part out of the image in order to encourage the network to extract discriminative features from other object's parts. Thus, through the attention-guided data augmentation, our model can extract more discriminative features in multiple object's parts, which means the object can be seen better.

Another benefit of our attention-guided data augmentation is that we can accurately locate objects, which makes our model look at objects closer and refine the predictions. For each testing image, object category will be coarse-to-fine predicted. The model first predicts object's region and coarse-stage probability of category from the raw image. Subsequently, the object's region is enlarged and the fine-stage probability is predicted.

In summary, the main contributions of this work are:
\begin{enumerate}
    \item We propose Weakly Supervised Attention Learning to generate attention maps to represent the spatial distribution of discriminative object's parts, and extract a sequential local features to solve the fine-grained visual classification problem.
    \item Based on attention maps, we propose attention-guided data augmentation to improve the efficiency of data augmentation, including attention cropping and attention dropping. Attention cropping randomly crops and resizes one of the attention part to enhance the local feature representation. Attention dropping randomly erases one of the attention region out of the image in order to encourage the model to extract the feature from multiple discriminative parts.
    \item We utilize attention maps to accurately locate the whole object and enlarge it to further improve the classification accuracy.

    Experiments in common fine-grained visual classification datasets demonstrates our data augmentation can significantly improve the accuracy of fine-grained classification and object localization, which surpasses the state-of-the-art methods and baselines.
\end{enumerate}

The rest of the paper is organized as follows. We first review the related works, including data augmentation and fine-grained visual classification in Section~\ref{sec:related_work}, then describe the proposed Weakly Supervised Data Augmentation Network (WS-DAN) in Section~\ref{sec:approach}. In Section~\ref{sec:experiments}, comprehensive experiments are conducted to demonstrate the effectiveness of WS-DAN. Finally, conclusion is drawn in Section~\ref{sec:conclusion}.


\section{Related Works}
\label{sec:related_work}
In this section, we review the related works of fine-grained image classification, data augmentation and weakly supervised localization.

\subsection{Fine-grained Visual Classification}
A variety of methods have been developed to distinguish different fine-grained categories. Convolutional Neural Networks were proposed to solve the large scale image classification problem. However, according to the experiments results in Table~\ref{tab:cub}, these basic models can only achieve the moderate performance, because it is relatively difficult for them to focus on the subtle differences of object's parts without special design.

To focus on the local features, many methods rely on the annotations of parts location or attribute. Part R-CNN~\cite{Part-rcnn}  extended R-CNN~\cite{rcnn} to detect objects and localize their parts under a geometric prior, then predicted a fine-grained category from a pose-normalized representation.  Lin \etal proposed a feedback-control framework Deep LAC~\cite{deeplac} to back-propagate alignment and classification errors to localization; they also proposed a valve linkage function (VLF) to connect the localization and classification modules.

To reduce aditional location labeling cost, methods that only require image-level annotation draw more attention. Different feature pooling methods have been proposed. Lin \etal proposed bilinear pooling~\cite{bcnn} and improved bilinear pooling~\cite{improved_bcnn}, where two features are combined at each location using the outer product, which considers their pairwise interactions. MPN-COV~\cite{mpn-cov} improved second-order pooling by matrix square and achieved the state-of-the-art accuracy.

Spatial Transformer Network (ST-CNN)~\cite{stn} aims to achieve accurate classification performance by first learning a proper geometric transformation and align the image before classifying. The method can also locate several object's parts at the same time. Fu \etal proposed Recurrent Attention CNN (RA-CNN)~\cite{ra-cnn} to recurrently predict the location of one attention area and extract the corresponding feature, while the method only focuses on one local part, so they combine three scale feature, i.e. three object's parts, to predict the final category.
To generate multiple attention locations at the same time, Zheng \etal proposed Multi-Attention CNN (MA-CNN)~\cite{ma-cnn}, which simultaneously locates several body parts. And they proposed channel grouping loss to generate multiple parts by clustering. However, the number of object's parts is limited (2 or 4), which might constrain their accuracy. In this paper, we propose bilinear attention pooling which combines attention layers with featur layers, whose number of attention regions is more easily to be increased and improve the classification accuracy as shown in Table~\ref{tab:num_parts}.

Also, metric learning has been introduced in FGVC task. Sun \etal proposed Multiple Attention Multiple Class (MAMC) loss ~\cite{mamc} that pulls positive features closer to the anchor, while pushes negative features away. Dubey \etal proposed PC~\cite{pairwise_confusion} to combine the pairwise confusion loss with cross entropy loss to learn features with greater generalization, thereby preventing overfitting. In our model, attention regularization loss is proposed to regular the attention regions and corresponding local features, which improves the identities of object's parts and the classification accuracy.

\subsection{Data Augmentation}
Our data augmentation method focuses on image's spatial augmentation. Before our work, random spatial image augmentation methods~\cite{max-drop, cutout, hide-and-seek}, such as image cropping and image dropping, have been proposed and proved to be effective in improving the robustness of deep models. Max-drop~\cite{max-drop} aims to remove the maximally activated features to encourage the network to consider the less prominent features. The drawback is that Max-drop can only remove one discriminative region of each image, which limits their performance. Cutout~\cite{cutout} and Hide-and-Seek~\cite{hide-and-seek} improve the robustness of CNNs by randomly masking many square regions out of training images. However, a great many of the erased regions are unrelavant background, or the whole object might be erased out, especially for small objects.

Random data augmentation suffers from low efficiency and generating much uncontrolled noisy data. To overcome these issue, a few methods have been proposed to take data distribution into consideration, which is more effective than a random data augmentation. Cubuk~\etal proposed AutoAugmentation~\cite{auto_autment} to create a search space of data augmentation policies. It can automatically design a specific policy so as to obtain state-of-the-art validation accuracy for target dataset. Peng~\etal proposed Adversarial Data Augmentation~\cite{ada} to jointly optimize data augmentation and deep model. They designed an augmentation network to online generate hard data and improve the robustness of the deep model. However, their data-specific augmentation is significantly complicated than random augmentation. Our attention-guided data augmentation is more simple and can generate part-level augmentation to boost the performance.

\subsection{Weakly Supervised Learning for Localization}
Weakly supervised learning is an umbrella term that covers a variety of studies that attempt to construct predictive models by learning with weak supervision ~\cite{review:wsl}. Accurately locating the object or its parts only by image-level supervision is very challenging. Early works~\cite{box_predict,Zhang2016TopDownNA} usually generate class-specific localization maps by Global Average Pooling (GAP) ~\cite{nin}. The activation area can reflect the location of an object. However, training by softmax cross entropy loss usually leads the model to pay attention to the most discriminative location, whose output bounding box just covers part of the object. To locate the whole object. Singh \etal ~\cite{hide-and-seek} randomly hides the patches of input images so as to force the network to find other discriminative parts. However, the process is inefficient for the lack of high-level guidance. Zhang \etal proposed Adversarial Complementary Learning (ACoL) ~\cite{acol} approach to discover entire objects by training two adversary complementary classifiers, which can locate different object's parts and discover the complementary regions that belong to the same object. Nevertheless, there are only two complementary regions in their implementation, which limits accuracy.  Our attention-guided data augmentation encourages the model to pay attention to multiple object's parts, extract more discriminative features and achieve significantly performance in object localizing, as shown in Table~\ref{tab:localiztion_error}.

\section{Approach}
\label{sec:approach}
\begin{figure*}[t]
    \centering
    \begin{subfigure}{0.60\textwidth}
        \centering
        \includegraphics[width=\linewidth]{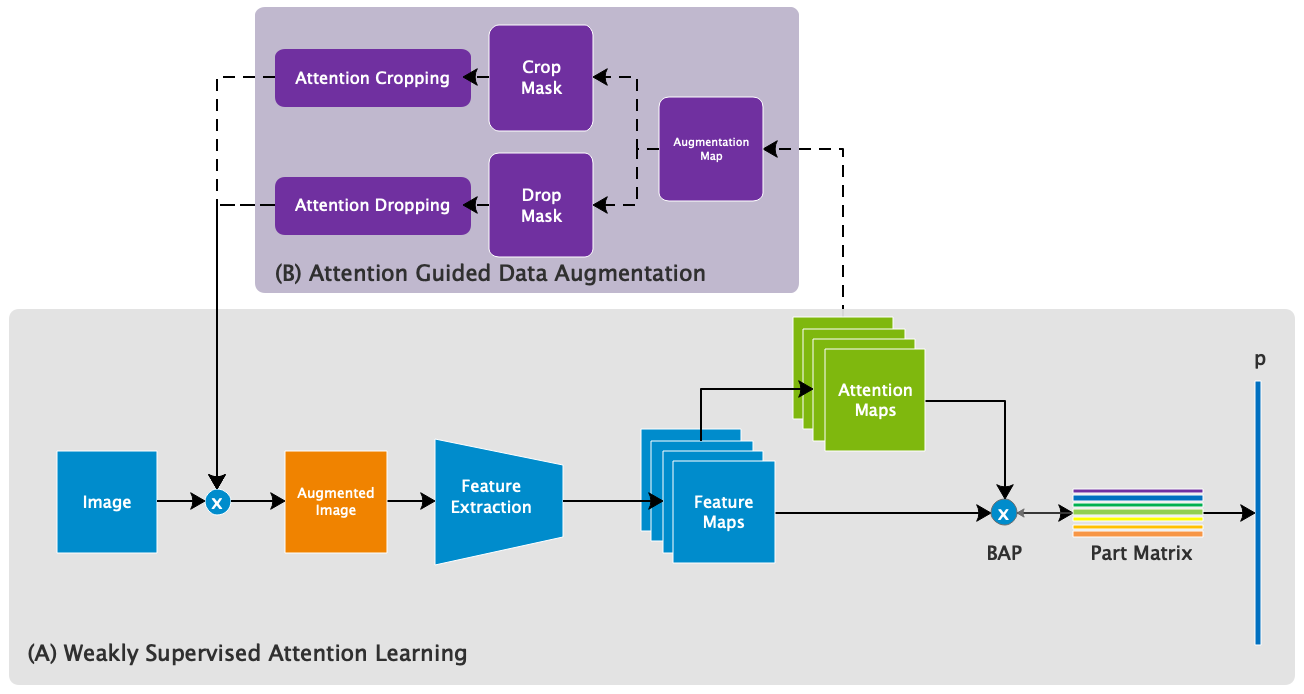}
        \caption{Training process. (A) Weakly Supervised Attention Learning. For each training image, attention maps will be generated to represent the object's discriminative parts by weakly supervised attention learning. (B) Attention-Guided Data Augmentation. One attention map is randomly selected to augment this image, including attention cropping and attention dropping. Finally, the raw and augmented data will be trained as input data.}
        \label{fig:training_process}
    \end{subfigure}

    \begin{subfigure}{0.60\textwidth}
        \centering
        \includegraphics[width=\linewidth]{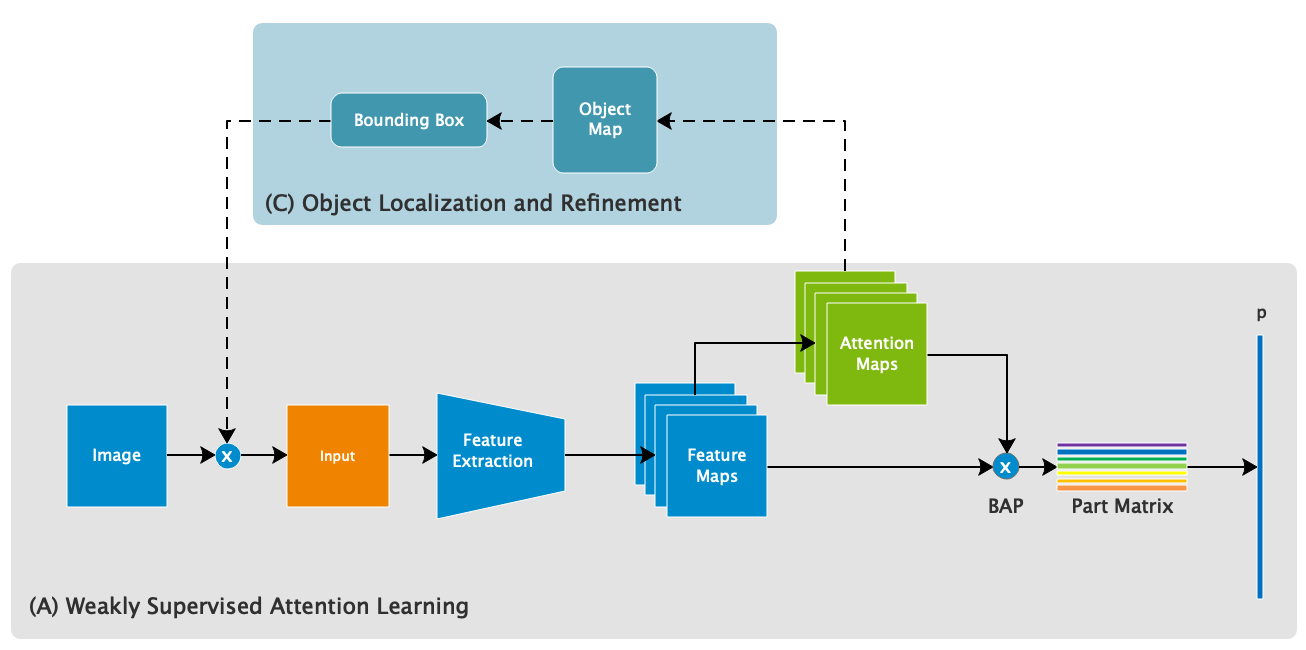}
        \caption{Testing process. Firstly, object's categories probability and attention maps will be outputed from raw image by (A). Secondly, object will be located according to (C) and then be enlarged to refine the categories probability. Finally, the above two probabilities will be combined as the final prediction.}
        \label{fig:testing_process}
    \end{subfigure}
    \caption{Overview training and testing process of Weakly Supservised Data Augmentation Network.}
    \label{fig:weakly_supervised_image_augmentation}
\end{figure*}

In this section, we describe the proposed WS-DAN in detail, including weakly supervised attention learning, attention-guided data augmentation and object localization \& refinement. The overview structure of proposed WS-DAN is illustrated in Fig~\ref{fig:weakly_supervised_image_augmentation}.

\subsection{Weakly Supervised Attention Learning}
\subsubsection{Spatial Representation}
We first predict the parts' regions of objects. During training and testing, the object's location annotation (\eg bounding boxes or keypoints) is not available. In our method, we adopt weakly supervised learning to predict objects' location distribution only by their category annotations.

We extract the feature of image $I$ by CNN and denote $F\in R^{H\times W \times N}$ as feature maps, where $H,W$ and $C$ represents feature layer's height, width and the number of channels respectively. The distributions of objects' parts are represented by Attention Maps $A\in R^{H\times W \times M}$ which is obtained from $F$ by
\begin{equation}
    A = f(F) =\bigcup_{k=1}^{M} A_k
\end{equation}
 where $f(\cdot)$ is a convolutional function. $A_k \in R^{H\times W}$ represents one of the object's part or visual pattern, such as the head of a bird, the wheel of a car or the wing of an aircraft. $M$ is the number of attention maps. Atteniton maps will be utilized to augment training data in Section~\ref{subsec:augmentation}.

 We propose regions of object's parts by attention maps rather than SS~\cite{ss} or RPN~\cite{faster-rcnn} since the former is more flexible and can be more easily trained end-to-end in FGVC task.

\subsubsection{Bilinear Attention Pooling}
After representing object's parts by attention maps $A$, inspired by Bilinear Pooling~\cite{bcnn} that aggregates feature representation from two-stream network layers, we propose Bilinear Attention Pooling (BAP) to extract features from these parts. We element-wise multiply feature maps $F$ by each attention map $A_k$ in order to generate $M$ part feature maps $F_k$, as shown in Equ~\ref{equ:attention_pooling}.

\begin{figure}[h]
    \begin{center}
        \includegraphics[width=1.0\linewidth]{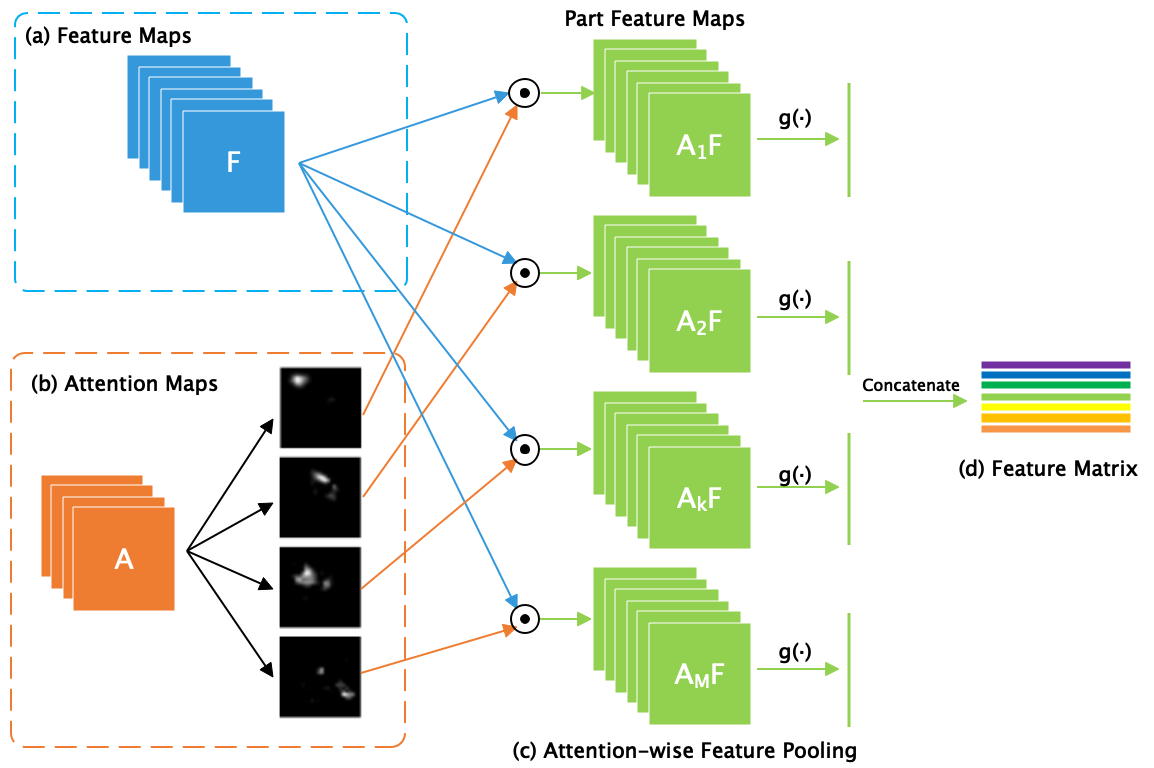}
    \end{center}
    \caption{\small The process of Bilinear Attention Pooling. The network backbone(\eg Inception v3~\cite{inception}) first generates feature maps (a) and attention maps (b) respectively . Each attention map represents a specific object's part; (c) Part feature maps are generated by element-wise multiplying each attention map with feature maps. Then, part features are extracted by convolutional or pooling operation; (d)The final feature matrix consists of all these part features.}
    \label{fig:attention_pooling}
\end{figure}

\begin{equation}
\begin{aligned}
F_k = A_k \odot F \quad(k = 1,2,...,M)
\end{aligned}
\label{equ:attention_pooling}
\end{equation}
where $\odot$ denotes element-wise multiplication for two tensors.

Then, we further extract discriminative local feature by additional feature extraction function $g(\cdot)$, such as Global Average Pooling (GAP), Global Maximum Pooling (GMP) or convolutions, in order to obtain $k_{th}$ attention feature $f_k\in R^{1 \times N}$,
\begin{equation}
f_k = g(F_k)
\end{equation}

Object's feature is represented by part feature matrix $P \in R^{M \times N} $ which is stacked by these part features $f_k$. Let $\Gamma(A, F)$ indicates bilinear attention pooling between attention maps $A$ and feature maps $F$. It can be represented in Equ~\ref{equ:label_matrix},
\begin{equation}
P= \Gamma(A, F)
= \begin{pmatrix} g (a_1 \odot F) \\ g(a_2 \odot F) \\ ... \\ g(a_M \odot F)  \end{pmatrix}
= \begin{pmatrix}f_1 \\ f_2 \\ ... \\ f_M \end{pmatrix}
\label{equ:label_matrix}
\end{equation}

\subsubsection{Attention Regularization}
For each fine-grained category, we expect that attention map $A_k$ can represent the same $k_{th}$ object's part.
Inspired by center loss~\cite{center_loss} proposed to solve face recognition problem, we propose attention regularization loss to weakly supervised the attention learning process. We penalize the variances of features that belong to the same object's part, which means that part feature $f_k$ will get close to the a global feature center $c_k\in R^{1 \times N}$ and attention map $A_k$ will be activated in the same $k_{th}$ object's part. The loss function can be represented by $L_A$ in Equ~\ref{equ:attention_center_loss},
\begin{equation}
L_{A} = \sum_{k=1}^M \| f_k - c_k \|_2^2
\label{equ:attention_center_loss}
\end{equation}
where $c_k$ is its part's feature center.  $c_k$ can be initialized from zero and updated by moving average,
\begin{equation}
c_k \gets c_k + \beta (f_k - c_k)
\end{equation}
where $\beta$ controls the update rate of $c_k$, and attention regularization loss only applies for raw images.


\subsection{Attention-guided Data Augmentation}
\label{subsec:augmentation}
\subsubsection{Augmentation Map}
Random data augmentation is low efficient, especially when the size of the object is small, and suffers from introducing a high percentage of background noises. With attention maps, data can be more efficiently augmented. For each training image, we randomly choose one of its attention map $A_k$ to guide the data augmentation process, and normalize it as $k_{th}$ Augmentation Map $A^*_k\in R^{H\times W}$.
\begin{equation}
  \begin{array}{l}
    A^*_k = \dfrac{A_k - \min(A_k)}{\max(A_k) - \min(A_k)} \\
  \end{array}
    \label{equ:normalized_map}
\end{equation}

\subsubsection{Attention Cropping}
With augmentation map $A^*_k$, we can zoom into this part's region and extract more detailed local features. Specifically, we first obtain the Crop Mask $C_k$ from $A^*_k$ by setting element $A^*_k(i, j)$ which is greater than threshold $\theta_c\in [0, 1]$ to $1$, and others to $0$, as represented in Equ~\ref{equ:attention_cropping}.
\begin{equation}
  \begin{array}{l}
        C_k(i, j)=\begin{cases}
            1, & \text{if $A^*_k(i, j) > \theta_c $}\\
            0, & \text{otherwise}.
          \end{cases} \\
  \end{array}
    \label{equ:attention_cropping}
\end{equation}

We then find a bounding box $B_k$ that can cover the whole selected positive region of $C_k$ and enlarge this region from raw image as the augmented input data, As illustrated in Fig~\ref{fig:attention_augmentation}. Since the scale of object's part increases, object can be seen better for extracting more fine-grained features.

\subsubsection{Attention Dropping}
Attention regularization loss supervises each attention map $A_k\in R^{H \times W}$ to represent the same $k_{th}$ object's part, while different attention maps might focus on the similar object's part. To encourage attention maps represent multiple discriminative object's parts, we propose attention dropping. Specifically, we obtain attention Drop Mask $D_k$ by setting element $A^*_k(i, j)$ which is greater than threshold $\theta_d\in [0, 1]$ to $0$, and others to $1$, as shown in Equ~\ref{equ:attention_dropping},
\begin{equation}
  \begin{array}{l}
        D_k(i, j)=\begin{cases}
            0, & \text{if $A^*_k(i, j) > \theta_d $}\\
            1, & \text{otherwise}.
          \end{cases} \\
  \end{array}
    \label{equ:attention_dropping}
\end{equation}
The $k_{th}$ part region will be dropped by masking image $I$ with $D_k$. The augmented image is illustrated in Fig~\ref{fig:attention_augmentation}.

Since $k_{th}$ object's part is eliminated from the image, the network will be encouraged to propose other discriminative parts, which means the object can also be seen better: the robustness of classification and the accuracy of localization will be improved, as shown in Table~\ref{tab:components}.


\subsection{Object Localization and Refinement}
The attention-guided data augmentation can also contributes to previous weakly supervised attention learning process, which means the location of object can be more accurately predicted. In the testing process, after the model outputs the coarse-stage classification result and corresponding attention maps for the raw image, we can predict the whole region of the object and enlarge it to predict fine-grained result by the same network model. Object Map $A_m$ that indicates the location of object is calculated by Equ~\ref{equ:mean_map}.
\begin{equation}
A_m = \dfrac{1}{M} \sum_{k=1}^{M} A_k
\label{equ:mean_map}
\end{equation}
The final classification result is averaged by the coarse-grained prediction and fine-grained prediction. The detailed process of Coarse-to-Fine prediction is described as Algorithm~\ref{alg:testing_process},
\begin{algorithm}[h]
    \begin{algorithmic}[1]
        \REQUIRE Trained WS-DAN model W
        \REQUIRE Raw image $I_r$
        \STATE Predict coarse-grained probability $p_1$: $p_1 = W(I_r)$ and output attention maps $A$;
        \STATE Calculate object map $A_m$ of $A$ by Equ.~\ref{equ:mean_map};
        \STATE Obtain the bounding box $B$ of object from $A_m$;
        \STATE Enlarge the region in $B$ as $I_o$;
        \STATE Predict fine-grained probability $p_2$:  $p_2 = W(I_o)$;
        \STATE Calculate the final prediction: $p = (p_1 + p_2) / 2$;
        \RETURN $p$
    \end{algorithmic}
    \caption{Object Localization and Refinement}
    \label{alg:testing_process}
\end{algorithm}

\section{Experiments}
\label{sec:experiments}
In this section, we show comprehensive experiments to verify the effectiveness of WS-DAN. Firstly, we explore the contribution of each proposed module. Then we compare our model with the state-of-the-art methods on four publicly available fine-grained visual classification datasets. Following this, we perform additional experiments to demonstrate the effect of the number of attention maps.
\subsection{Datasets and Experiments Settings}
\paragraph{Datasets} We compare our method with the state-of-the-art on four FGVC datasets, including CUB-200-2011~\cite{CUB_200_2011}, FGVC-Aircraft~\cite{fgvc_aircraft}, Stanford Cars~\cite{Stanford_car} and Stanford Dog~\cite{stanford_dog}. Specific information of each dataset is shown in Table~\ref{tab:dataset}.

\begin{table}[h]
    \begin{center}
        \scriptsize
        \begin{tabular}{c|c|c|c|c}
            \hline
            Dataset & Object & \#Category & \#Training & \#Testing\\
            \hline
            CUB-200-2011 & Bird & 200 & 5,994 & 5,794  \\
            \hline
            FGVC-Aircraft & Aircraft & 100 & 6,667 & 3,333\\
            \hline
            Stanford Cars & Car & 196 & 8,144 & 8,041 \\
            \hline
            Stanford Dogs & Dog & 120 & 12,000 & 8,580 \\
            \hline
        \end{tabular}
    \end{center}
    \caption {Introduction to four common fine-grained visual classification datasets.}
    \label{tab:dataset}
\end{table}

\subsection{Implement Details}
In the following experiments, we adopt Inception v3~\cite{inception} as the backbone and choose \textit{Mix6e} layer as feature maps. Attention maps are obtained by $1 \times 1$ Convolutional kernel. We adopt GAP as the feature pooling function $g(\cdot)$. The attention cropping and dropping threshold $\theta_c$ and $\theta_d$ are both set to $0.5$.

We train the models using Stochastic Gradient Descent (SGD) with the momentum of $0.9$, epoch number of $80$, weight decay of $0.00001$, and a mini-batch size of $16$ on a P100 GPU. The initial learning rate is set to $0.001$, with exponential decay of $0.9$ after every $2$ epochs. The code will be released in the near feature.

\subsection{Accuracy Contribution}
As described above, our WS-DAN mainly consists of four components, including Weakly Supervised Attention Learning, Attention Cropping, Attention Dropping, and Object Localization and Refinement (Loc.\& Refinement). We perform an experiment in CUB-200-2011 dataset and demonstrate that each component is effective to improve the accuracy, as shown in Table~\ref{tab:components}.
\begin{table}[h]
    \begin{center}
        \scriptsize
        \begin{tabular}{c|c|c|c|c}
            \hline
            \makecell{Attention \\ Learning} & \makecell{Attention \\ Cropping} & \makecell{Attention \\ Dropping} & \makecell{Loc.\&Refinement} & Accuracy(\%)\\
            \hline
             & & & & 83.7 \\
             \checkmark & & & & 86.4\\
             \hline
             \checkmark & \checkmark & & & 87.8\\
             \checkmark & & \checkmark & & 87.4\\
             \checkmark & \checkmark & \checkmark & & 88.4\\
             \hline
             \checkmark & \checkmark & \checkmark & \checkmark & \textbf{89.4}\\
            \hline
        \end{tabular}
    \end{center}
    \caption {Contribution of proposed components and their combinations.}
    \label{tab:components}
\end{table}

\subsection{Comparision with Random Data Augmentation}
We conduct this experiment to demonstrate the effectiveness of attention-guided data augmentation compared with random data augmentation. The baseline method is our Weakly Supervised Attention Learning model. For clarity, we evaluate the performance only for the first stage. In this experiment, we evaluate the accuracy of object localization by Mean Intersection-over-Union (mIoU). Higher score of mIoU means more accurately in locating the object. The results are shown in Table~\ref{tab:random_augmentation}. We can see that attention-guided data augmentation is more efficient than random data augmentation.

\begin{table}[h]
    \begin{center}
        \scriptsize
        \begin{tabular}{c|c|c}
            \hline
            Data Augmentation & Accuracy(\%) & mIoU(\%)\\
            \hline
            Baseline & 86.4 & 38.4 \\
            \hline
            Random Cropping & 86.8 & 40.0 \\
            Attention Cropping & \textbf{87.8} & \textbf{58.7} \\
            \hline
            Random Dropping~\cite{cutout} & 86.7 & 45.8 \\
            Attention Dropping & \textbf{87.4} & \textbf{59.1}\\
            \hline
            \makecell{Attention Cropping\\ + Attention Dropping} & \textbf{88.4} & \textbf{62.0}\\
            \hline
        \end{tabular}
    \end{center}
    \caption {Comparison with random data augmentation in CUB-200-2011 dataset.}
    \label{tab:random_augmentation}
\end{table}

\subsection{Comparison with Stage-of-the-Art Methods}
\paragraph{Fine-grained Classification Results}
We compare our method with state-of-the-art methods on above mentioned fine-grained classification datasets. The results are respectively shown in Table~\ref{tab:cub}, Table~\ref{tab:fgvc}, Table~\ref{tab:car} and Table~\ref{tab:dog}. It can be seen that our WS-DAN achieves the state-of-art accuracy on all these fine-grained datasets. Espetially, we significantly improve the accuracy compared with the backbone Inception v3.

\begin{table}[h]
	\begin{center}
		\scriptsize
		\begin{tabular}{c|c}
			\hline
			Methods  & Accuracy(\%)\\
			\hline
			VGG-19~\cite{vgg} & 77.8 \\
			ResNet-101~\cite{resnet} & 83.5\\
			Inception-V3~\cite{inception} & 83.7\\
      \hline
			B-CNN~\cite{bcnn} & 84.1\\
			ST-CNN~\cite{stn} & 84.1\\
			PDFR~\cite{PDFR} &  84.5\\
			RA-CNN~\cite{ra-cnn} &  85.4\\
      GP-256~\cite{gp} & 85.8\\
			MA-CNN~\cite{ma-cnn} & 86.5\\
      MAMC~\cite{mamc} & 86.5 \\
      PC~\cite{pairwise_confusion} & 86.9 \\
      DFL-CNN~\cite{dfl-cnn} & 87.4 \\
      NTS-Net~\cite{nes-net} & 87.5\\
      MPN-COV~\cite{mpn-cov} & 88.7\\
      \hline
      \textbf{WS-DAN} & \textbf{89.4} \\
			\hline
		\end{tabular}
	\end{center}
	\caption {Comparison with state-of-the-art methods on CUB-200-2011 testing dataset. }
	\label{tab:cub}
\end{table}

\begin{table}[h]
	\begin{center}
		\scriptsize
		\begin{tabular}{c|c}
			\hline
			Methods  & Accuracy(\%)\\
			\hline
      VGG-19~\cite{vgg} & 80.5 \\
			ResNet-101~\cite{resnet} & 87.2\\
			Inception-V3~\cite{inception} & 87.4 \\
			\hline
			B-CNN~\cite{bcnn} & 84.1\\
			RA-CNN~\cite{ra-cnn}  &  88.4  \\
      PC~\cite{pairwise_confusion} & 89.2 \\
      GP-256~\cite{gp} & 89.8\\
			MA-CNN~\cite{ma-cnn} & 89.9  \\
      MPN-COV~\cite{mpn-cov} & 91.4\\
      NTS-Net~\cite{nes-net} & 91.4\\
      DFL-CNN~\cite{dfl-cnn} & 92.0 \\
			\hline
			\textbf{WS-DAN}  & \textbf{93.0}\\
			\hline
		\end{tabular}
	\end{center}
	\caption {Comparison with state-of-the-art methods on FGVC-Aircraft testing dataset. }
	\label{tab:fgvc}
\end{table}

\begin{table}[h]
	\begin{center}
		\scriptsize
		\begin{tabular}{c|c}
			\hline
			Methods & Accuracy(\%)\\
			\hline
      VGG-19~\cite{vgg} & 85.7\\
			ResNet-101~\cite{resnet} & 91.2\\
			Inception-V3~\cite{inception} & 90.8 \\
			\hline
			RA-CNN~\cite{ra-cnn} & 92.5\\
			MA-CNN~\cite{ma-cnn} & 92.8  \\
      GP-256~\cite{gp} & 92.8\\
      PC~\cite{pairwise_confusion} & 92.9 \\
      MAMC~\cite{mamc} & 93.0 \\
      MPN-COV~\cite{mpn-cov} & 93.3\\
      DFL-CNN~\cite{dfl-cnn} & 93.8 \\
      NTS-Net~\cite{nes-net} & 93.9\\
      \hline
			\textbf{WS-DAN} & \textbf{94.5}\\
			\hline
		\end{tabular}
	\end{center}
	\caption {Comparison with state-of-the-art methods on Stanford Cars testing dataset. }
	\label{tab:car}
\end{table}

\begin{table}[h]
    \begin{center}
        \scriptsize
        \begin{tabular}{c|c}
            \hline
            Method & Accuracy(\%)\\
            \hline
            VGG-19~\cite{vgg} & 76.7\\
            ResNet-101~\cite{resnet} & 85.8\\
            Inception-V3~\cite{inception} & 88.9 \\
            \hline
            NAC~\cite{nac} & 68.6\\
            PC~\cite{pairwise_confusion} & 83.8 \\
            FCAN~\cite{fcan} & 84.2 \\
            MAMC ~\cite{mamc} & 85.2\\
            RA-CNN~\cite{ra-cnn} & 87.3\\
            \hline
            \textbf{WS-DAN} & \textbf{92.2}\\
            \hline
        \end{tabular}
    \end{center}
    \caption {Comparison with state-of-the-art methods on Stanford Dog testing dataset. }
    \label{tab:dog}
\end{table}

\paragraph{Object Localization Results}
 The recent method ACoL~\cite{acol} provided the performances of image-based object localization on CUB-200-2011 bird dataset. To compare with it, we evaluate our method on the same dataset with the same metric, \ie calculating the localization error (failure percentage) of the images whose bounding boxes have less 50\% IoU with the ground truth. Since most of the images of FGVC-Aircrafts and Stanford Cars datasets occupy the whole space, we only evaluate the localization error rate in CUB-200-2011 and Stanford Dogs datasets.

 The experimental results are shown in Table~\ref{tab:localiztion_error}. Our model surpasses the state-of-the-art methods and baselines by a large margin.


\begin{table}[h]
	\begin{center}
		\scriptsize
		\begin{tabular}{c|c|c}
			\hline
			Method & CUB-200-2011(\%) & Stanford Dogs(\%)\\
			\hline
			GoogLeNet  &  59.0 & 30.7 \\
      VGGnet-ACoL~\cite{acol} &  54.1 & - \\
      ResNet-101~\cite{resnet} & 42.1 & 29.6 \\
      Inception V3~\cite{densenet} & 40.8 & 28.8 \\
      \hline
			\textbf{WS-DAN} & \textbf{18.3} & \textbf{19.2}\\
			\hline
		\end{tabular}
	\end{center}
	\caption {Object Localization error rate on CUB-200-2011 and Stanford Dogs datasets. The localization error rate of our model is significantly lower than other methods.}
	\label{tab:localiztion_error}
\end{table}

\subsection{Effect of Number of Attention Maps}
More object's parts usually contribute to the better performance. Similar conclusion has been made in MA-CNN~\cite{ma-cnn} and ST-CNN~\cite{stn}. We perform this experiment about the effectiveness of the number of attention maps $M$. Table~\ref{tab:num_parts} shows that with the increasing of $M$, the classification accuracy also rises. When $M$ reaches to around 32, the performance gradually becomes stable, and the final accuracy reaches to 89.4\%. Our feature pooling model makes it easy to set an arbitrary number of object's parts. We can achieve a more accurate result by increasing the number of attention maps.

\begin{table}[h]
    \begin{center}
        \scriptsize
        \begin{tabular}{c|c}
            \hline
            \# Attention Maps & Accuracy(\%)\\
            \hline
            4 & 86.6 \\
            8 & 87.7 \\
            16 & 88.3 \\
            32 & 89.4 \\
            64  & 89.4 \\
            \hline
        \end{tabular}
    \end{center}
    \caption {The effect of number of attention maps evaluated on CUB-200-2011 dataset. }
    \label{tab:num_parts}
\end{table}


%
%

\subsection{Visualization of Augmented Data}
\begin{figure}[h]
    \centering
    \begin{subfigure}{0.5\textwidth}
        \centering
        \includegraphics[width=\linewidth]{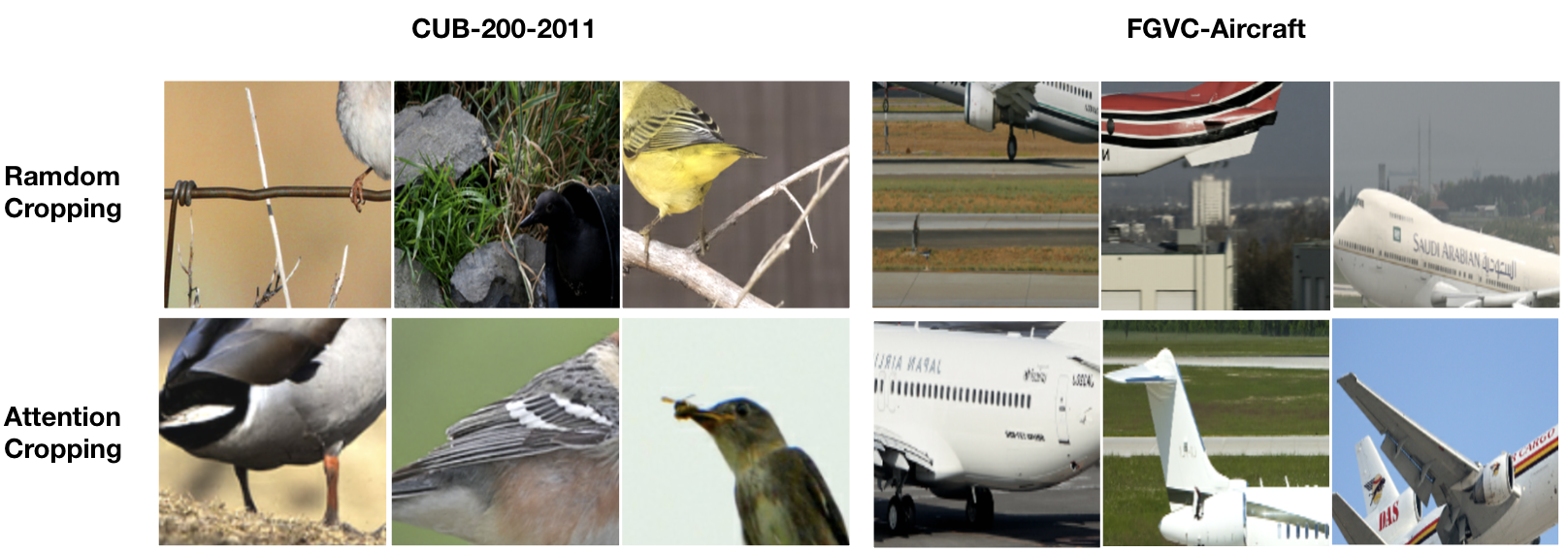}
        \caption{Comparision between Attention Cropping and Random Cropping. Random cropping is very likely to include a high percentage of background as input image, while attention cropping knows exactly where to crop to see better.}
        \label{fig:dan}
    \end{subfigure}

    \begin{subfigure}{0.5\textwidth}
        \centering
        \includegraphics[width=\linewidth]{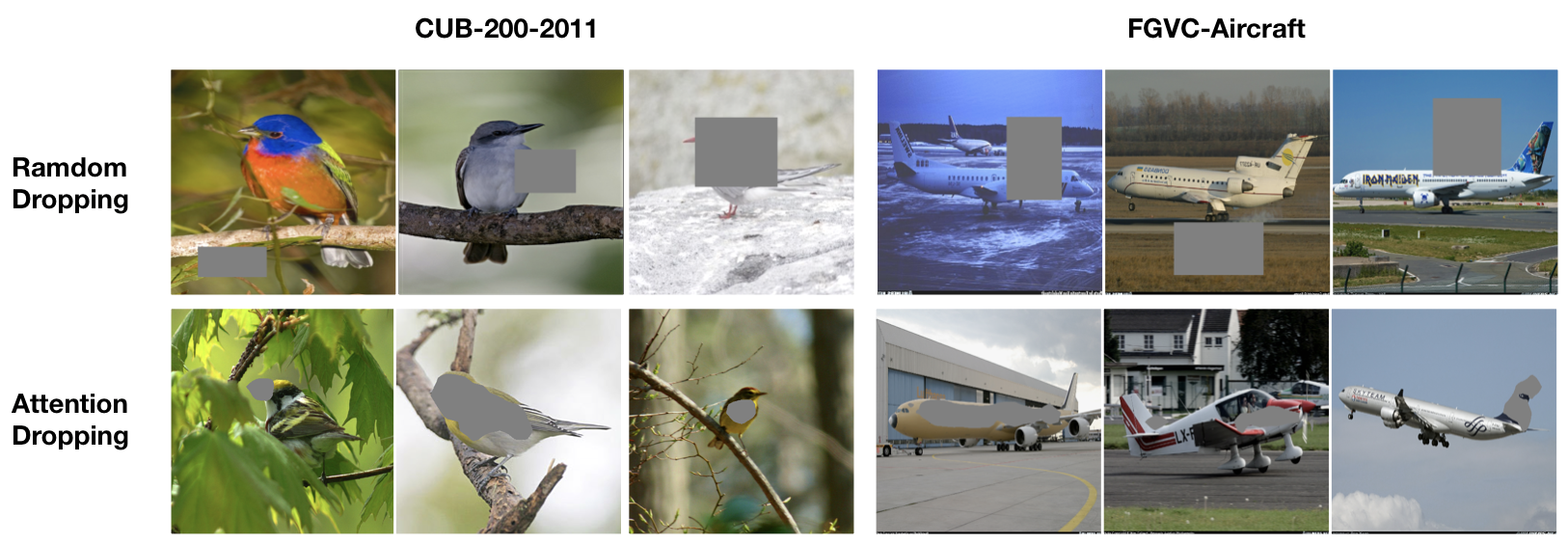}
        \caption{Comparision between Attention Dropping and Random Dropping. Random dropping might erase the whole object out of the image or just erase background. Attention dropping is more efficient for erasing the discriminative object parts and promoting multiple attention.}
        \label{fig:ws_dan}
    \end{subfigure}
    \caption{Visualization of augmented images of random and attention-guided data augmentation.}
    \label{fig:augmented_data}
\end{figure}

In Fig~\ref{fig:augmented_data}, we visualize the augmented images by random data augmentation and attention-guided data augmentation in CUB-200-2011 and FGVC-Aircraft datasets. Intuitively, random data augmentation introduces much background into the training data. Attention-guided data augmentation is more efficient in cropping or dropping because of the guidance of location of objects' parts.

\section{Conclusion}
\label{sec:conclusion}
In this paper, we propose Weakly Supervised Data Attention Network to significantly. We combine Weakly Supervised Learning (WSL) with Data Augmentation (DA). WSL provides objects' spatial distribution for DA and DA encourages the attention learning process of WSL. They benefit from each other and promote the model to extract more discriminative image feature and from multiple local regions, which ensures WS-DAN to surpass the state-of-the-art methods.

{\small
\bibliographystyle{ieee}
\bibliography{iccv}
}

\end{document}